\documentclass{article} 
\usepackage{iclr2019_conference,times}
\usepackage{graphicx}

\usepackage{amsmath,amsfonts,bm}

\def\eqref#1{equation~\ref{#1}}

\def\1{\bm{1}}

\DeclareMathAlphabet{\mathsfit}{\encodingdefault}{\sfdefault}{m}{sl}
\SetMathAlphabet{\mathsfit}{bold}{\encodingdefault}{\sfdefault}{bx}{n}

\newcommand{\E}{\mathbb{E}}

\usepackage{algorithmicx}
\usepackage{algorithm}
\usepackage{algorithm,algpseudocode}
\usepackage{authblk}
\usepackage{hyperref}
\usepackage{url}
\usepackage{leftidx}

\usepackage{array}
\usepackage{booktabs}

\usepackage{color}

\title{ Structured Content Preservation for Unsupervised Text Style Transfer }
\author[*]{\bf{Youzhi Tian}}
\author[**]{\bf{Zhiting Hu}}
\author[*]{\bf{Zhou Yu}}
\affil[*]{University of California, Davis}
\affil[**]{Carnegie Mellon University}

\begin{document}

\maketitle
%
%
\begin{abstract}
Text style transfer aims to modify the style of a sentence while keeping its content unchanged. Recent style transfer systems often fail to faithfully preserve the content after changing the style. This paper proposes a structured content preserving model that leverages linguistic information in the structured fine-grained supervisions to better preserve the style-independent content \footnote{Henceforth, we refer to style-independent content as content, for simplicity.} during style transfer. In particular, we achieve the goal by devising rich model objectives based on both the sentence's lexical information and a language model that conditions on content. The resulting model therefore is encouraged to retain the semantic meaning of the target sentences. We perform extensive experiments that compare our model to other existing approaches in the tasks of sentiment and political slant transfer.\footnote{All the code and data used in the experiments have been released to facilitate reproducibility at https://github.com/YouzhiTian/Structured-Content-Preservation-for-Unsupervised-Text-Style-Transfer.} Our model achieves significant improvement in terms of both content preservation and style transfer in automatic and human evaluation.
\end{abstract}

\section{Introduction}
Text style transfer is an important task in designing sophisticated and controllable natural language generation (NLG) systems. The goal of this task is to convert a sentence from one style (e.g., negative sentiment) to another (e.g., positive sentiment), while preserving the style-independent content (e.g., the name of the food being discussed). 
Typically, it is difficult to find parallel data with different styles. So we must learn to disentangle the representations of the style from the content. However, it is impossible to separate the two components by simply adding or dropping certain words. Content and style are deeply fused in every sentence. 
 
Previous work on unsupervised text style transfer, such as \citet{2017arXiv170300955H,2017arXiv170509655S}, proposes an encoder-decoder architecture with style discriminators to learn disentangled representations. The encoder takes a sentence as an input and generates a style-independent content representation. The decoder then takes the content representation and the style representation to generate the transferred sentence.  
However, there is no guarantee that the decoder will only change the style-dependent content, as these methods do not have any constraints to preserve the content. 

We propose a new text generative model to preserve the content while transferring the style of a sentence at the same time. 
We first use an attentional auto-encoder (AE) and a binary style classifier to generate sentence with the target style. Then we use the distance between POS information of the input sentence and output sentence as an error signal to the generator. Since we use the positive-negative Yelp review data \citep{2017arXiv170509655S} and the political slant data as the benchmark to test our algorithm. Most of the content of these reviews and comments are captured by the nouns of the sentence. Therefore, we enforce the decoder to generate sentences that have similar nouns. So in our implementation, the POS information simply reduces to noun information. For other style transfer tasks, different POS tags, such as verbs, may capture the content better. Then the model needs to adapt to use other tags instead of the nouns for the constraints. Besides these explicit constraints, we also add an additional constraint, a language model conditioned on both the nouns and the style representations, to enforce the output sentence to be fluent and contain desired nouns.

We evaluate our model on both the sentiment modification and the political slant transfer tasks. Results show that our approach has better content preservation and a higher accuracy of transferred style compared to previous models in both automatic and human evaluation metrics.

\section{Related work}
Recently, there are many new models designed for non-parallel text style transfer. Our structured content preserving model is closely related to several previous work describe below:

\textbf{Delete, Retrieve and Generate} According to the observation that text styles are often marked by distinctive phrases (e.g., ``a great fun”), \citet{2018arXiv180406437L} extracts content words by deleting phrases associated with the sentence’s original style, retrieves new phrases associated with the target style, and uses a neural model to smoothly combine them into a final output. However, in some cases \citep{TACL732}, content and style cannot be so cleanly separated only using phrase boundaries. 

\textbf{Back-Translation} \citet{2017arXiv171100043L,2017arXiv171011041A} propose sophisticated methods for unsupervised machine translation. \citet{2016arXiv161005461R} shows that author characteristics are significantly obfuscated by both manual and automatic machine translation. So these methods could in principle be used as style transfer. \citet{2018arXiv180409000P} first uses back-translation to rephrase the sentence to get rid of the original style. Then, they generate a sentence based on the content using a separate style-specific generator. However, after back-translation, the generated sentence does not contain the detailed content information in the input sentence.   

\textbf{Adversarial Training} Some work \citep{2017arXiv170509655S,2017arXiv170604223Z,2017arXiv171106861F,2017arXiv171109395M} uses adversarial training to separate style and content. For example, \citet{2017arXiv171106861F} proposes two models: multi-decoder and style-embedding. Both of them learn a representation for the input sentence that only contains the content information. Then the multi-decoder model uses different decoders, one for each style, to generate sentences in the corresponding style. The style-embedding model, in contrast, learns style embeddings in addition to the content representations. \citet{2017arXiv170509655S} also encodes the source sentence into a vector, but the discriminator utilizes the hidden states of the RNN decoder instead. When applying adversarial training to unsupervised text style transfer, the content encoder aims to fool the style discriminator by removing style information from the content embedding. However, empirically it is often easy to fool the discriminator without actually removing the style information. Besides, the non-differentiability of discrete word tokens makes the generator difficult to optimize. Hence, most models attempt to use REINFORCE \citep{Sutton:1999:PGM:3009657.3009806} to fine tune trained models \citep{2016arXiv160905473Y,2017arXiv170106547L} or use professor forcing method \citep{2017arXiv170106547L} to transfer the style.

\textbf{Language Model as Discriminator} \citet{2018arXiv180511749Y} first uses a target domain language model to provide token-level feedback during the learning process. In their approach, they train the language model as a discriminator to assign a high probability to real sentences, replacing the more conventional binary classifier. This method is different from ours in that our language model is conditioned on the content. So the content conditional language model can enforce the decoder to preserve the content.

\textbf{Toward Controlled Generation of Text} \citet{2017arXiv170300955H} is most relevant to our work. Their model aims at generating sentences with controllable styles by learning disentangled latent representations \citep{2016arXiv160603657C}. It builds on variational auto-encoders (VAEs) and uses independency constraints to enforce styles reliably inferred back from generated sentences. Our model adds more constraints to preserve the style-independent content by using POS information preservation and a content-conditional language model.



\section{Structured Content Preserving Model}
\begin{figure*}[tp]
\centering
\includegraphics[width=1.0\textwidth]{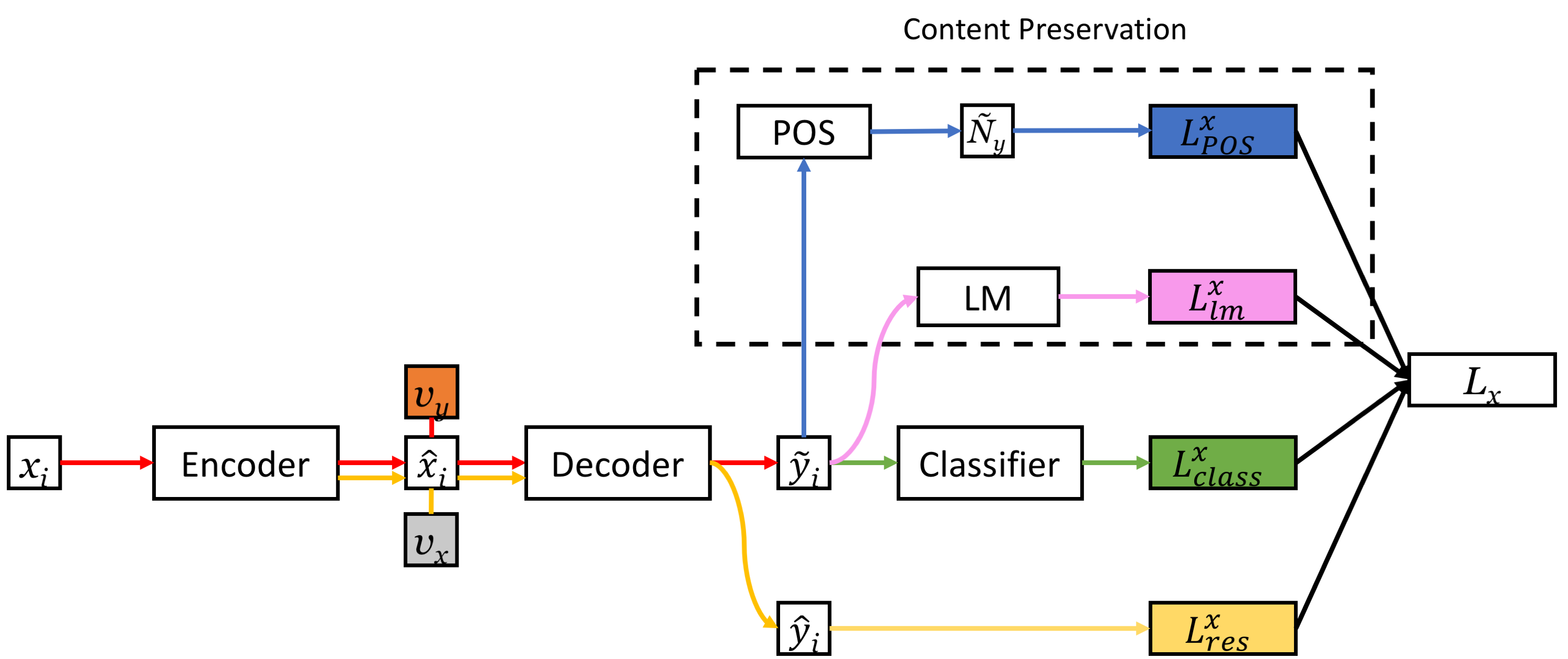}
\caption{Architecture of our proposed model. $x_{i}$ denotes the input sentence. $\hat{x}_i$ denotes the style-independent content representation. $v_{x}$ and $v_{y}$ denote the different style representations. ${\tilde{y}}_i$ denotes the generated sentence. $\tilde{N}_y$ denotes the noun set extracted from $\tilde{y}_i$. $L_{res}^x$, $L_{POS}^x$, $L_{class}^x$, $L_{lm}^x$ and $L_x$ denote the loss function. The red line shows the generation process. The yellow, blue, green and pink lines illustrate the loss function formulation. }\label{Models}
\end{figure*}

The overall model architecture is shown in Figure~\ref{Models}. 
We develop a new neural generative model which integrates an attentional auto-encoder, a style classifier, a POS information preservation constraint and a language model that conditions on content \citep{2018arXiv180511749Y}. We will discuss each component in details below.

\subsection{Attentional Auto-encoder}
Assume there are two non-parallel text datasets $X$ = \{$x_1$, $x_2$,\dots, $x_m$\} and $Y$ = \{$y_1$, $y_2$,\dots, $y_m$\} with two different styles $v_x$ and $v_y$, respectively.  We let $v_x$ be the positive sentiment style and $v_y$ be the negative sentiment style. Our target is to generate sentence of the desired style while preserving the content of the input sentence.
Assume the style representation $v$ is sampled from a prior $p(v)$ and the content representation $z$ is sampled from $p(z)$. So the sentence $x$ is generated from the conditional distribution $p(x|z,v)$ and our goal is to learn:
\begin{equation}
    p(y|x) = \int_{z_x} p(y|z_x,v_y)q(z_x|x,v_x) dz_x
\end{equation}
The above equation is the encoder and decoder framework to generate sentences. One unsupervised method is to apply the attentional auto-encoder model. 
We first encode the sentence $x_i$ or $y_i$ to obtain its style-independent content representation $\hat{x}_i = E(x_i,v_x)$ or $\hat{y}_i = E(y_i,v_y)$. Then we use an attentional decoder $G$ to generate sentences conditioned on $\hat{x}_i$ or $\hat{y}_i$ and $v_x$ or $v_y$. So we can get the reconstruction loss $\mathcal{L}_{res}^x$:
\begin{equation}
    \mathcal{L}_{res}(\theta_{E}|\theta_{G}) = \E_{\text x \sim \text X}[-\log p_G(x|\hat{x}_i,v_x)]+\E_{\text y \sim \text Y}[-\log p_G(y|\hat{y}_i,v_y)]
\end{equation}
After switching the style representation, we can get the generated sentence ${\bf{ \tilde{y}_i}}= \{{\tilde{y}}_1,{\tilde{y}}_2,\dots,{\tilde{y}}_M\}$:
\begin{equation}
    {\bf{\tilde{y}}} = G(\hat{x}_i,v_y) = p_G({\bf{\tilde{y}}}|\hat{x}_i,v_y) = \prod_{m=1}^{M}p(\tilde{y}_m|{\bf{\tilde{y}}}^{<m},\hat{x}_i,v_y)
\end{equation}
where ${\bf{\tilde{y}}}^{<m}$ indicates the tokens preceding $\tilde{y}_m$ and $M$ denotes the number of tokens in ${\bf{ \tilde{y}_i}}$ . 

\subsection{Style Classifier}
Since the attentional auto-encoder cannot fully enforce the desired style, we deploy a style classifier to predict the given style. We feed the result of the prediction back to the encoder-decoder model. The generator is then trained to generate sentences that maximize the accuracy of the style classifier. 



However, it is impossible to propagate gradients from the style classifier through the discrete samples. We thus resort to a Gumbel-softmax \citep{2016arXiv161101144J} distribution as continuous approximation of the sample. For each step in the generator, let $q$ be a categorical distribution with probabilities $\pi_1, \pi_2, \dots, \pi_c$. Samples from $q$ can be approximated by:
\begin{equation}
u_i = \frac{e^{(\log(\pi_i)+g_i)/\tau}}{\sum_{j=1}^{c} e^{(\log(\pi_j)+g_j)/\tau}} \label{approxi}
\end{equation}
$g_i$ or $g_j$ denotes independent samples from Gumbel(0, 1). The temperature $\tau$ is set to $\tau$ $\rightarrow 0$ as training proceeds, yielding increasingly peaked distributions that finally emulate the discrete case. The approximation replaces the sampled token ${\tilde{y}}_i$ (represented as a one-hot vector) at each step with ${\bf{u}}$.
Then the probability vector ${\bf{u}}$ is used as the input to the classifier. With the continuous approximation, we use the following loss to improve the generator:
\begin{equation}
    \mathcal{L}_{class}(\theta_{G}) = \E_{p(\hat{x}_i)p(v_y)}[-\log p_C(\tilde{G}_{\tau}(\hat{x}_i,v_y))] + \E_{p(\hat{y}_i)p(v_x)}[-\log p_C(\tilde{G}_{\tau}(\hat{y}_i,v_x))]
\end{equation}
Where the $\tilde{G}_{\tau}(\hat{x}_i,v_y)$ and $\tilde{G}_{\tau}(\hat{y}_i,v_x))$ denote the resulting ``soft'' generated sentence and $p_C$ denotes the loss in the style classifier.

In most cases, the classifier is unstable and insufficient to preserve content. We propose to use POS information preservation and a language model to preserve content. 
\subsection{POS Preservation Constraints}
We propose to preserve the content based on the POS tag information. Since we focus on Yep review and political comments and their content are mostly captured by nouns, we only consider the noun tags of the POS information. We use a POS tagger to extract nouns from both input sentences and generated sentences. We denote the extracted noun set of the input as $N_x$ and the noun set in the output as ${\tilde{N}_y}$, where $N_x = \{n_1,n_2,\dots,n_p\}$ and ${\tilde{N}_y} = \{\tilde{n}_1,\tilde{n}_2,\dots,\tilde{n}_m\}$. Then we embed these nouns using pretrained GloVe \citep{pennington2014glove} to measure the semantic similarity between the input and output sentences. We embed each noun token in $N_x$ and ${\tilde{N}_y}$: 
\begin{equation}\label{embedding}
{E_x} = \{{\bf e_1},{\bf e_2},\dots,{\bf e_p}\} = Embedder(N_x)
\end{equation}
\begin{equation}
{\tilde{E}_y} = \{{\bf {\tilde{e}_1}},{\bf {\tilde{e}_2}},\dots,{\bf {\tilde{e}_m}}\} = Embedder({\tilde{N}_y})
\end{equation}
Where $E_x$ or $\tilde{E}_y$ denotes the embedding sets and ${\bf {e_1}}$ or ${\bf {\tilde{e}_1}}$ denotes the embedding of the nouns. However, the structure of the output sentence can be different from the input. Therefore, we need to match corresponding noun pairs between $N_x$ and ${\tilde{N}_y}$. 
Assume that the corresponding noun of $n_i$ in $N_x$ is ${\tilde{n}_j}$ in ${\tilde{N}_y}$ when ${\tilde{n}_j}$ is most similar to $n_i$ among all the nouns in ${\tilde{N}_y}$. We add up the cosine similarity between the two word embeddings of all the corresponding noun pairs and normalize it by the number of nouns in $N_x$ as a basic loss. 

In practice, the number of nouns in $N_x$ may not be equal to the ones in ${\tilde{N}_y}$, which means the generated sentence contains more or less nouns than the original sentence. So calculating the similarity of the two sentences can be challenging. Therefore, we include a weight  $\gamma$ to handle those cases to ensure a fair comparison. When the number of nouns in $N_x$ is equal to that in ${\tilde{N}_y}$, $\gamma$ is set to 1. When the number of nouns in $N_x$ is different to the number in ${\tilde{N}_y}$, $\gamma$ is set to be the absolute difference normalized by the number of nouns in $N_x$. So the POS distance is:
\begin{equation}\label{nn}
\mathcal{L}_{POS}(\theta_{G}) =\gamma \sum_{i=1}^{min(c_N^x,c_{\tilde{N}}^y)}d_i (d_i \in D = \{d_1,d_2,\dots,d_n\} )
\end{equation}
Where $\gamma = 1 + (max(c_N^x,c_{\tilde{N}}^y)-min(c_N^x,c_{\tilde{N}}^y))/c_N^x$ , the distance set denotes as D, the number of nouns in $N_x$ denotes as $c_N^x$ and the number of nouns in ${\tilde{N}_y}$ denotes as $c_{\tilde{N}}^y$. 

The POS distance can be viewed as an overall evaluation score of the output sentence. If the output does not contain semantically similar nouns of the input, the overall evaluation score will be high. The generator thus can be enhanced to generate sentence with desired nouns. In this way, the style-independent content can be better preserved.

\subsection{Content Conditional Language Model}
Besides the sentence POS information preservation, we use a content conditional language model to enforce the output sentence to contain desired nouns and at the same time to make the output sentence more fluent. Instead of comparing the difference between $N_x$ and $\tilde{N}_y$ directly, we only use $N_x$ to give noun information to the generator implicitly. We combine the average of $E_x$ (Equation~\ref{embedding}) with the style label as the hidden state $h_{lm}$ and send the hidden state to the language model. Thus the language model is conditioned on both the nouns and the style. Then we use the discrete samples from the input $x_i$ and the hidden state $h_{lm}$ to train the language model:
\begin{equation}\label{lm}
    \mathcal{L}_{lm}(\theta_{lm}) = \E_{\text x \sim \text X}[-\log p_{lm}(x_i|h_{lm})]
\end{equation}
In order to use standard back-propagation to train the generator, we use the continuous approximation in Equation~\ref {approxi}. We denote the continuous approximation of the output of the decoder as $\tilde{p}^y$ = $\{\tilde{p}_t^y\}_{t=1}^T$, which is a sequence of probability vectors. For each step we feed $\tilde{p}_t^y$ to the language model using the weighted average of the embedding $W\tilde{p}_t^y$, then we get the output from the language model which is a probability distribution over the vocabulary of the next word $\hat{p}^y_{t+1}$. Finally, the language model loss is the perplexity of the generated samples evaluated by the language model becomes:
\begin{equation}
\mathcal{L}_{lm}(\theta_{G}) = \E_{\text x \sim \text X,\tilde{p}^y \sim p_G(\tilde{y}|z_x,v_y)}[\sum_{t=1}^T (\tilde{p}_t^y)^T \log \hat{p}^y_t]
\end{equation}
If a sentence does not contain semantically similar nouns from the input, it will have a high perplexity under a language model trained on desired nouns. A language model can assign a probability to each token, thus providing more information on which word is to blame for overall high perplexity.

Therefore, the training objective for our structured content preserving model is to minimize the four types constraints together: the reconstruction in the attentional auto-encoder, the style classifier, the POS information and the language model:
\begin{equation}
\mathcal{L} = \mathcal{L}_{res} + \alpha\mathcal{L}_{class} + \beta\mathcal{L}_{POS} + \eta\mathcal{L}_{lm}
\end{equation}
The model training process consists of three steps. First, we train the language model according to Equation~\ref{lm}. Then, we train the classifier. Finally, we minimize the reconstruction loss, classification loss, POS loss and perplexity of the language model together.


\section{Experiments and Results}
We experiment on two types of different transformations: sentiment and political slant, to verify the effectiveness of our model. Compared with a broad set of exiting work, our model achieves significant improvement in terms of both content preservation and style transfer. 
\subsection{Accuracy of POS Tagger}
Before the experiment, we test the accuracy of the POS tagger. We first select 100 sentences from the test set. Then for each sentence, we annotate nouns manually and compare with those extracted by POS Tagger. Result shows the accuracy (\%) of the POS Tagger is 91.4. So the error of the POS tagger has little effect on the experiment results.

\subsection{Sentiment Manipulation}
We first investigate whether our model can preserve the sentiment-independent content better while transferring the sentiment of the sentence. We compare our model with \citet{2017arXiv170300955H,2017arXiv170509655S,2017arXiv171106861F,2018arXiv180409000P,2018arXiv180406437L}. We also evaluate our model without language model supervision in an ablation study.
\subsubsection{Dataset and Experimental Setting}
We use the Yelp review dataset. The dataset contains almost 250K negative sentences and 380K positive sentences, of which 70\% are used for training, 10\% are used for evaluation and the remaining 20\% are used as test set. Sentences of length more than 15 are filtered out. We keep all words that appear more than five times in the training set and get a vocabulary size of about 10k. All words appearing less than five times are replaced with a $<$UNK$>$ token. At the beginning of each sentence, we add a $<$BOS$>$ token and at the end of each sentence, we add a $<$EOS$>$ token. We use $<$PAD$>$ token to pad sentences that have less than 15 tokens. Appendix \ref{Con} shows the model configurations.

\begin{table}[t]
\begin{center}
\begin{tabular}{rllll}
\cmidrule[\heavyrulewidth]{1-5}
\multicolumn{1}{r}{\bf Model}  &\multicolumn{1}{l}{\bf Accuracy(\%)}  &\multicolumn{1}{l}{\bf BLEU} &\multicolumn{1}{l}{\bf BLEU(human)}
&\multicolumn{1}{l}{\bf POS distance}
\\ 
\cmidrule{1-5}
\cite{2017arXiv170300955H} &86.7 &58.4 &22.3 &1.038  \\ 
\cmidrule{1-5}
\cite{2017arXiv170509655S}    &73.9 &20.7 &7.8 &3.954\\ 
\cmidrule{1-5}
\cite{2018arXiv180409000P} &91.2 &2.8 &2.0 &5.923 \\ 
\cmidrule{1-5}
\cite{2017arXiv171106861F}: \\ 
StyleEmbedding    &8.1 &\bf 67.4 &19.2 &1.734 \\ 
MultiDecoder &46.9 &40.1 &12.9 &3.451 \\ 
\cmidrule{1-5}
\cite{2018arXiv180406437L}:   \\
Delete &85.5 &34.6 &13.4 &1.816 \\ 
Retrieval &\bf 97.9 &2.6 &1.5 &5.694 \\ 
Template &80.1 &57.4 &20.5 &0.867 \\
DeleteAndRetrieval &88.9 &36.8 &14.7 &2.053 \\ \cmidrule{1-5}
Our model (without lm) &90.1 &60.4 &23.7 &0.748 \\
Our model (without POS) &91.0 &58.7 &23.2 &0.650\\
Our model  &92.7 &63.3 &\bf24.9 &\bf0.569 \\ 
\cmidrule[\heavyrulewidth]{1-5}
\end{tabular}
\caption{Our model and baselines performance. The BLEU (human) is calculated using the 1000 human annotated sentences as ground truth from \citep{2018arXiv180406437L}. The POS distance denotes the noun difference between the original and transferred sentences. The smaller the POS distance, the better the performance. lm here represents language model.}
\label{Result1}
\end{center}
\end{table}

\subsubsection{Automatic Evaluation Results}
Following previous work \citep{2017arXiv170300955H,2017arXiv170509655S,2018arXiv180406437L}, we compute automatic evaluation metrics: accuracy and BLUE score. Besides these two basic metrics, we also use the POS information distance (Equation~\ref{nn}) as new metric to test the model's content preserving ability.

We first use a style classifier to assess whether the generated sentence has the desired style. The classifier is based on CNN and trained on the same training data. We define the accuracy as the fraction of outputs classified as having the targeted style. However, simply evaluating the sentiment of the sentences is not enough, since the model can generate collapsed sentences such as a single word ``good". We also measure the POS information distance and BLEU score of the transferred sentences against the original sentences. Besides these metrics, \citet{2018arXiv180406437L} provides 1000 human annotated sentences. These sentences  are manually generated by human that have the desired style and the preserved content. We use them as the ground truth and use BLEU score of transferred sentences against these human annotated sentences as an important metric to evaluate all the models.

We report the results in Table~\ref{Result1}. The Retrieval method in \citet{2018arXiv180406437L} has the highest accuracy but has very low score in both BLEU against original sentences and BLEU against human transferred sentences. Such result is not surprising, because they directly retrieve the most similar sentences from the other corpus. However, there is no mechanism to guarantee that the corpus with the target style contains the sentence that has similar content of the original sentence. The StyleEmbedding model in \citet{2017arXiv171106861F} has the highest BLEU score while its accuracy is very low. Such result is not surprising because the model does not have any constraints on the generated style. The results of these two models show that a good model should perform well on both the accuracy and the BLEU score. Comparing our model with other baselines, we can see that our model outperforms them in both aspects: getting higher accuracy and preserving the content better. This demonstrates the effectiveness of our structure content preserving model. 

Having both the POS and the language model constraints might look repetitive at first. Therefore, we experiment the model without the POS and language model constraints respectively. We report the results in Table~\ref{Result1}. The BLEU(human) gets improved from 23.2 to 24.9 with POS constraint. Because POS constraint mainly focuses on preserving the word semantics. The result also shows that the accuracy rises from 90.1 to 92.7 with language model constraint. It is not surprising that the language model helps the generator to generate sentence more fluently and select the appropriate words conditions on the content information. The result indicates having both POS and language model constraints are useful. 

Table~\ref{Result1} also shows that in most cases, if the model has a high BLEU score, it will usually have a small POS information distance. It seems that it is not necessary to use the POS infomration. However, when we compare our model with StyleEmbedding \citep{2017arXiv171106861F}, we can find that StyleEmbedding has both higher BLEU score and a higher POS information distance than ours. The BLEU (human) shows that our model preserves the content better than StyleEmbedding. BLEU(Human) and the POS information distance is correlated.  It is not surprising that the POS information distance measures the semantic similarity of corresponding nouns instead of counting the number of same tokens like BLEU score. For example in a good transfer case that transfers ``The meal is great!'' into ``The food is terrible'', the BLEU score of this generated sentence is low while the score of POS information distance is not. Therefore, The POS information distance is better than BLEU score to evaluate the ability of the model in preserving contents.


\subsubsection{Human Evaluation Results}
While the automatic evaluation metrics provide some indication of transfer quality, it does not capture all the aspects of the style transfer task. Therefore, we also perform human evaluations. \citet{2018arXiv180406437L} provide 500 sentences with positive sentiment and 500 sentences with negative sentiment. All the 1000 sentences are randomly selected from the test set in Yelp review dataset. They also provide the generated sentences from their model and other models \citep{2017arXiv170509655S,2017arXiv171106861F}. We first use these 1000 sentences as input and generate sentences from our model and other two baselines \citep{2017arXiv170300955H,2018arXiv180409000P}. It is hard for users to make decisions when presented with multiple sentences at the same time. So instead of comparing all the models at the same time, we compare our model with all other models one by one. In each human evaluation, we randomly select 100 sentences from the set. Then for each original sentence, we present two corresponding outputs of our model and a baseline model in a random order. The annotator is then asked  ``Which sentence has an opposite sentiment of the original sentence and at the same time preserves the content of it?" They can choose: ``A", ``B" or ``the same". We also hired different annotators to rate each pair of models, so nobody will see the same sentence twice.

Table~\ref{Result2} shows the human evaluation results. Evaluators prefer our model than other models in general. Results in Table~\ref{Result1} show that the Retrieval model in \citet{2018arXiv180406437L} has the highest accuracy and StyleEmbedding model in \citet{2017arXiv171106861F} has the highest BLEU score. However, there are 72 sentences generated from our model that are better than those from these two models respectively. Sentences generated from our model have a higher overall quality.

\begin{table}[h]
\begin{center}
\begin{tabular}{rlll}
\cmidrule[\heavyrulewidth]{1-4}
\multicolumn{1}{r}{\bf Model}  &\multicolumn{1}{l}{\bf Our Model(\%)} &\multicolumn{1}{l}{\bf Other Model(\%)} & \multicolumn{1}{l}{\bf The Same(\%) }\\ 
\cmidrule{1-4}
\cite{2017arXiv170509655S}    &67 &6 &27\\
\cmidrule{1-4}
MultiDecoder    &85 &5 &10 \\ 
\cmidrule{1-4}
\cite{2017arXiv170300955H} &55 &17 &28\\
\cmidrule{1-4}
\cite{2018arXiv180406437L} &72 &28  &0 \\ 
\cmidrule{1-4}
styleEmbedding &72 &4 &24\\
\cmidrule{1-4}
\cite{2018arXiv180409000P} &56 &4 &40\\ 
\cmidrule[\heavyrulewidth]{1-4}
\end{tabular}
\caption{Our model is generally preferred over other models in human evaluation}
\label{Result2}
\end{center}
\end{table}

\begin{table}[h]
\begin{center}
\begin{tabular}{rlll}
\cmidrule[\heavyrulewidth]{1-2}
\multicolumn{1}{r}{\bf Category}  &\multicolumn{1}{l}{\bf Sentence}  
\\ 
\cmidrule{1-2}
Original &actually , just keep walking .  \\ 
Transferred    &actually , just keep walking .\\ 
\cmidrule{1-2}
Original &the service has always been wonderful .\\
Transferred &the service has not been terrible . \\
\cmidrule{1-2}
Original &the wait staff is extremely attractive and friendly !\\
Transferred &the wait staff is extremely attractive and rude !\\ 
 \cmidrule[\heavyrulewidth]{1-2}
\end{tabular}
\caption{Some bad examples generated by our model.}
\label{badsamples}
\end{center}
\end{table}

\subsubsection{Analysis}
We list some examples of transferred sentences in Table~\ref{samples} in Appendix. Both \citet{2017arXiv170300955H} and \citet{2017arXiv170509655S} change the meaning of the original sentences (e.g. ``the menudo here is perfect.'' $\rightarrow$ ``the terrible here is awkward.''). It shows that only using classifier as an error signal to the generator is unstable. Sometimes it changes the style-independent tokens into sentiment tokens to reach a high accuracy in the style classifier. \citet{2018arXiv180409000P} always change the detailed content information (e.g. ``the menudo here is perfect." $\rightarrow$ ``the fare is horrible''). \citet{2018arXiv180406437L} sometimes changes the structure of the original sentences (e.g. ``if you travel a lot do not stay at this hotel.'' $\rightarrow$ ``would highly recommend this place if you travel a lot at this hotel.''). 
Compared with other models, our model preserves the content better while changing the sentiment (e.g. ``the service was excellent and my hostess was very nice and helpful.'' $\rightarrow$ ``the service was terrible and my hostess was very disappointing and unhelpful.''). Because after adding constraints on POS tags, the output sentences preserve the content better. The content-conditional language model also feeds the noun information to the generator. Therefore, the source and generated sentence has semantically similar content. Our style classifier precision is high as well. Because after adding constraints to nouns, the generator can only change other style-dependent parts, such as adjectives in the sentence. Therefore, it is more likely to transfer the sentence to the desired style correctly. 

However, there are still some cases that our model cannot handle well. We show some examples in Table~\ref{badsamples}. If the sentiment of the sentences are not obvious (e.g. ``actually, just keep walking."), the generated sentences sometimes would be the same as the input sentences. In some cases, our transferred sentence has double negation (e.g. the service has not been terrible) which results in error in the style transferred. For sentences with complex sentence structures, such as conjunction, in some rare cases, we miss certain part of sentence in the conversation (e.g. ``the wait staff is extremely attractive and rude!")

\begin{table}[t]
\begin{center}
\begin{tabular}{rlll}
\cmidrule[\heavyrulewidth]{1-4}
\multicolumn{1}{r}{\bf Model}  &\multicolumn{1}{l}{\bf Accuracy(\%)} &\multicolumn{1}{l}{\bf BLEU} &\multicolumn{1}{l}{\bf POS distance}
\\ 
\cmidrule{1-4}
\cite{2018arXiv180409000P} &86.5 &7.38 &7.298  \\ 
\cmidrule{1-4}
\cite{2017arXiv170300955H} &90.7 &47.5 &3.524 \\
\cmidrule{1-4}
Our model    &\bf92.4 &\bf56.6 &\bf2.837 \\ 
 \cmidrule[\heavyrulewidth]{1-4}
\end{tabular}
\caption{Our proposed model outperforms other results on the political slant dataset.}\label{politicalResult}
\label{political}
\end{center}
\end{table}

\subsection{Political Slant Transfer}
We have demonstrated that the structured content preserving model can successfully preserve the content while transferring the sentiment of the sentence. To verify the robustness of our model, we also use it to transfer the political slant between democratic and republican party. Compared with \citet{style_transfer_acl18,2017arXiv170300955H}, our model still achieves improvement in terms of both content preservation and style transfer.
\subsubsection{Dataset and Experimental Setting}
The political dataset is comprised of top-level comments on Facebook posts from all 412 current members of the United States Senate and House who have public Facebook pages \citep{rtgender}. The data set contains 270K democratic sentences and 270K republican sentences and the preprocessing steps and the experiment configurations are the same as the previous experiment. 

\subsubsection{Automatic Evaluation Results}
We use three automatic metrics: accuracy, BLUE score and POS information distance to evaluate our model against \citet{style_transfer_acl18,2017arXiv170300955H}. We report the results in Table~\ref{politicalResult}. Our model outperforms the other two models in all three metrics. However, the BLEU score is lower than the score in the sentiment modification task (56.6 VS 63.3). This is because names of politicians (e.g. Donald Trump) are relevant to the political slant. The classifier therefore will enforce the generator to transfer these names to those with another political slant. However, the POS constraints will preserve these people names. Therefore, the two constraints have a conflict and make the generation results less ideal. However, our model can still retain high BLEU score and low POS distance if there are no explicit people names in the source sentences.


\section{Conclusion}
Based on the encoder-decoder framework, we develop a new neural generative model, names structured content preserving model. It integrates an attentional auto-encoder, a style classifier, a POS constraint and a content-conditional language model. Our model outperforms previous models in both the sentiment and political slant transfer tasks. Compared with other models, our model preserve the content better while changing the style of the original sentence. We also propose a new evaluation metric: POS distance, which measures the content preservation between the source and transferred sentence to evaluate the content preserving ability of the model.

\newpage
\bibliography{iclr2019_conference}
\bibliographystyle{iclr2019_conference}

\newpage
\appendix

\section{Model Configurations}
\label{Con}
We use similar model configuration to that of \citep{2017arXiv170300955H} for a fair comparison. The encoder and language model is one-layer GRU \citep{2014arXiv1412.3555C}. The decoder (generator) is an attentional \citep{2014arXiv1409.0473B} one-layer GRU. The word embedding size is 100 and GRU hidden size is 700. v is the style representation vector of size 200. The CNN classifier (i.e., discriminator) is trained with 128 filters and the kernel size is chosen from \{3,4,5\}. The parameters of the language model and classifier are not shared with parameters of other parts and are trained from scratch. We use a batch size of 128 with negative samples and positive samples. We use Adam \citep{2014arXiv1412.6980K} optimization algorithm to train the language model, classifier and the attentional auto-encoder and the learning rate is set to be the same. Hyper-parameters are selected based on the validation set. We use grid search to pick the best parameters. The learning rate is set to 5e-4. $\eta$, the weight of language model loss, is set to 0.5. $\alpha$, the weight of classifier loss, is set to 0.2. $\beta$, the weight of noun loss, is set to 0.1. Models are trained for a total of 10 epochs (pretrain one epoch). The best result is obtained after 4 epochs. We use an annealing strategy to set the temperature of $\tau$ of the Gumbel-softmax approximation. The initial value of $\tau$ is set to 1.0 and it decays by half every epoch (after pretrain) until reaching the minimum value of 0.001.

\begin{table}[p]
\begin{tabular}{rl}
\cmidrule[\heavyrulewidth]{1-2}
\multicolumn{1}{r}{\bf Model}  &\multicolumn{1}{l}{\bf Positive To Negative} \\
\cmidrule{1-2}
Original &the happy hour crowd here can be fun on occasion.\\ 
\cite{2017arXiv170300955H} &the shame hour crowd sucked can be sloppy on occasion.\\
\cite{2018arXiv180406437L} &crowd here can be ok on occasion.\\
\cite{2017arXiv170509655S} &the wait hour , won't be happy at home down.\\
\cite{2018arXiv180409000P} &the worst service at the food is the worst.\\
Our model(content preserving) &the unhappy hour crowd here can be disappointed on occasion.\\ 
\cmidrule{1-2}
Original &the menudo here is perfect. \\
\cite{2017arXiv170300955H} &the terrible here is awkward.\\
\cite{2018arXiv180406437L} &sadly the menudo here is inedible.\\
\cite{2017arXiv170509655S} &the family here is an understatement.\\
\cite{2018arXiv180409000P} &the fare is horrible. \\
Our model(content preserving) &the menudo here is nasty.\\ 
\cmidrule{1-2}
Original &the service was excellent and my hostess was very nice and helpful. \\
\cite{2017arXiv170300955H} &the awful was nasty and my hostess was very nasty and helpful.\\
\cite{2018arXiv180406437L} &what was too busy to my hostess than nothing to write trash.\\
\cite{2017arXiv170509655S} &the service was dirty and the office was very nice and helpful and.\\
\cite{2018arXiv180409000P} &the service is ok and the food wasn't even and they were halls.\\
Our model(content preserving) &the service was terrible and my hostess was very disappointing and unhelpful.\\ 
\cmidrule[\heavyrulewidth]{1-2}
\bf Model &\bf Negative To Positive \\ 
\cmidrule{1-2}
Original &if you travel a lot do not stay at this hotel. \\
\cite{2017arXiv170300955H} &remarkable you travel a lot do brilliant stay at this hotel.\\
\cite{2018arXiv180406437L} &would highly recommend this place if you travel a lot at this hotel. \\
\cite{2017arXiv170509655S} &if you use a lot , i stay at this hotel.\\
\cite{2018arXiv180409000P} &if you want i love this place in the valley.\\
Our model(content preserving) &if you travel a lot do always stay at this hotel.\\ 
\cmidrule{1-2}
Original &maria the manager is a horrible person.\\
\cite{2017arXiv170300955H} &maria the manager is a delicious person.\\
\cite{2018arXiv180406437L} &maria the manager is a great customer service person.\\
\cite{2017arXiv170509655S} &love the place is a wonderful guy.\\
\cite{2018arXiv180409000P} &flan is always a great experience.\\
Our model(content preserving) &maria the manager is a perfect person.\\ \hline
Original &avoid if at all possible. \\
\cite{2017arXiv170300955H} &impressed remarkable at all possible.\\
\cite{2018arXiv180406437L} &absolutely the best gyros at rei.\\
\cite{2017arXiv170509655S} &love god at all possible. \\
\cite{2018arXiv180409000P} &definitely recommend this place. \\
Our model &recommend if at all possible.\\ 
\cmidrule[\heavyrulewidth]{1-2}
\label{samples}
\end{tabular}
\caption{Sentiment transfer examples}
\label{samples}
\end{table}

\end{document}